\ifdefined\pdfminorversion
\fi
\documentclass[10pt,twocolumn,letterpaper]{article}
\usepackage{volucompose-preprint}

\graphicspath{{figures/}}
\newcommand{\method}{VoluCompose}
\newcommand{\tableformat}{\footnotesize\setlength{\tabcolsep}{3pt}\renewcommand{\arraystretch}{1.02}}
\title{VoluCompose: Constructing Compact Volumetric\\
Visual Representations for Medical Image Understanding}
\author{%
  Zanting Ye\affmark{1,2,\textdagger}\quad
  Yunrui Lin\affmark{1,\textdagger}\quad
  Shengyuan Liu\affmark{3,\textdagger}\quad
  Jing Hao\affmark{4}\quad
  Chen Hu\affmark{2}\quad
  Yefeng Zheng\affmark{2,*}\quad
  Lijun Lu\affmark{1,*}%
}
\paperaffiliations{%
  \affmark{1}\,School of Biomedical Engineering and Guangdong Provincial Key Laboratory of Medical Image Processing,\\
  Southern Medical University, Guangzhou 510515, China\\
  \affmark{2}\,Medical Artificial Intelligence Laboratory, Westlake University, Hangzhou 310030, China\\
  \affmark{3}\,Department of Electronic Engineering, The Chinese University of Hong Kong, Hong Kong SAR, China\\
  \affmark{4}\,Faculty of Dentistry, The University of Hong Kong, Hong Kong SAR, China%
}
\paperauthornote{%
  \affmark{\textdagger}\,Equal contribution\qquad
  \affmark{*}\,Correspondence:
  \href{mailto:zhengyefeng@westlake.edu.cn}{zhengyefeng@westlake.edu.cn};
  \href{mailto:ljlubme@gmail.com}{ljlubme@gmail.com}%
}
\date{}
\hypersetup{
  pdftitle={VoluCompose: Constructing Compact Volumetric Visual Representations for Medical Image Understanding},
  pdfauthor={Zanting Ye, Yunrui Lin, Shengyuan Liu, Jing Hao, Chen Hu, Yefeng Zheng, Lijun Lu},
  pdfkeywords={Volumetric image understanding; vision-language models; training-free token compression}
}

\begin{document}
\paperbibcontrol{VoluComposeBSTControl}
\maketitle

\begin{abstract}
Medical vision--language models can interpret volumetric images, but processing full volumes is costly because hundreds of slices produce long, redundant visual-token sequences. Existing compression methods often couple content selection with budget determination. We observe that neighboring slices share visual features and that attention concentration varies across visual encoders. We propose VoluCompose, a training-free, model-agnostic framework that separates what visual content to retain from how many tokens to allocate. Its hyperbolic token composition module groups contiguous slices into ordered support units and uses the Lorentz distance to account for feature magnitude and direction during token allocation, anchor selection, and refinement. Its capacity control module uses visual-encoder attention entropy to allocate more tokens when attention is more dispersed. We evaluate VoluCompose on two tasks across three datasets and four backbones. Across nine calibrated settings, VoluCompose attains the highest mean quality--retention hypervolume among the compared methods. It achieves a mean relative quality of 121.68\% against the uncompressed references while retaining 22.78\% of visual tokens on average. Compression remains robust even at a 10\% token budget. Across settings, the lowest ROUGE-L ratio to the uncompressed reference is 96.4\% for VoluCompose. The corresponding minima for the compression baselines range from 15.0\% to 70.5\%. Code is available at
\url{https://github.com/yezanting/VoluCompose}.
\end{abstract}

\begin{paperkeywords}
Volumetric image understanding, vision--language models, training-free token compression.
\end{paperkeywords}

\begin{figure}[!t]
\centering\includegraphics[width=0.94\columnwidth]{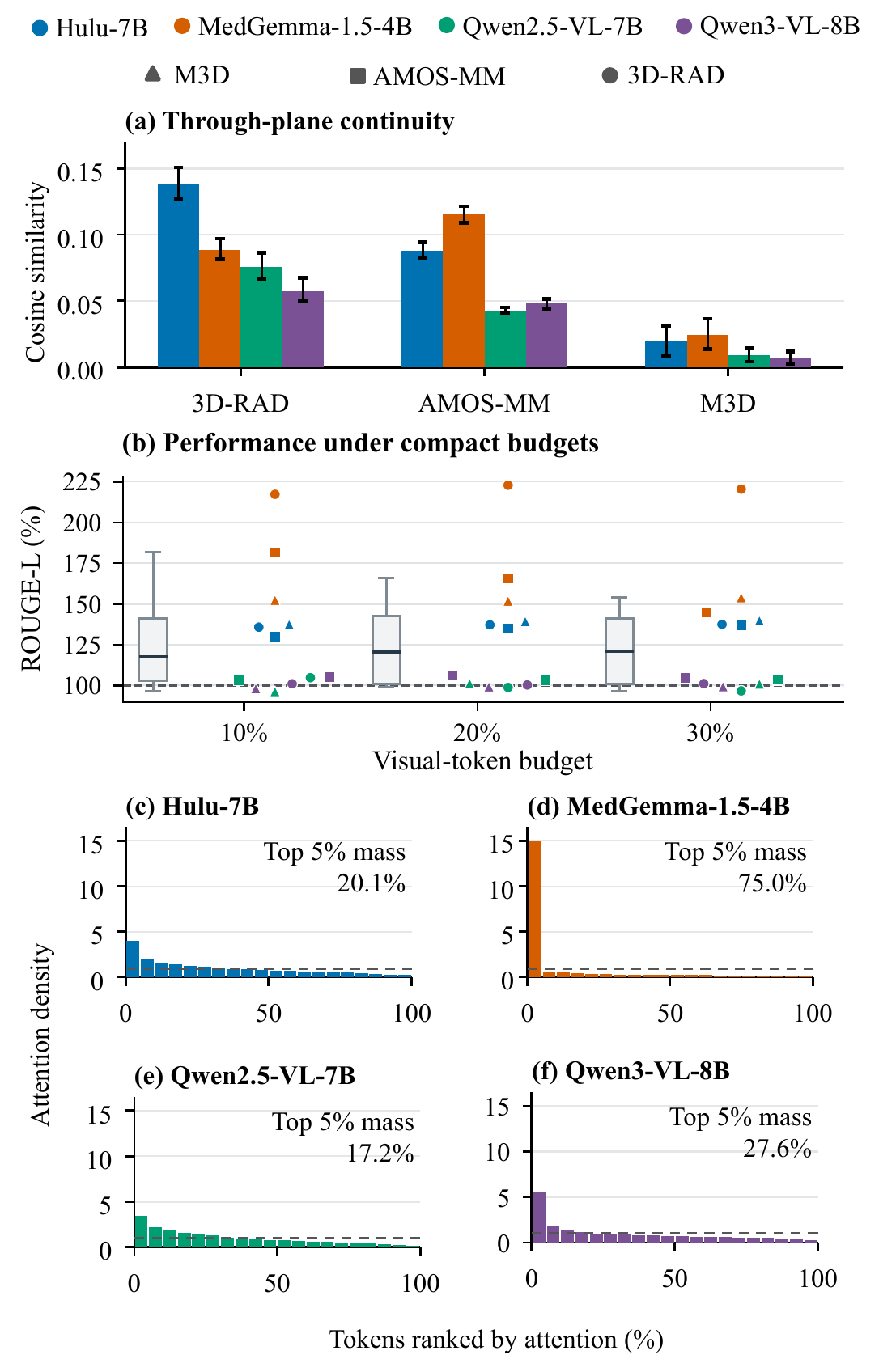}
\caption{Cross-slice feature recurrence, attention heterogeneity, and performance under compact token budgets. (a) Adjacent-minus-nonadjacent cosine similarity of slice features; bars show means with 95\% bootstrap intervals over 50 studies per dataset. (b) Setting-level ROUGE-L of \method{} relative to the uncompressed reference (100\%) under fixed nominal budgets (10\%, 20\%, and 30\%). (c)--(f) Ranked attention density averaged over 150 profiles per backbone; annotations give the attention mass in the top 5\% of tokens.}
\label{fig:motivation}
\end{figure}

\Needspace{4\baselineskip}
\section{Introduction}
\label{sec:introduction}

Vision--language models (VLMs) combine visual perception with language-based reasoning for medical image understanding~\cite{acosta2022,moor2023foundation,ye2026volumetric}. Recent models further extend this capability from individual images to multi-image, video, and volumetric inputs~\cite{qwen252025,qwen32025,xu2025lingshu,hulu2025,medgemma2026}. However, a full volume may contain hundreds of ordered slices, each producing a dense grid of visual tokens. Directly forwarding these tokens incurs quadratic self-attention computation during decoder prefill, increases key--value cache memory demand, and may exceed the available context budget~\cite{photon2026,orca2026}. Efficient volumetric understanding therefore requires a compact representation that preserves clinically relevant visual information with fewer tokens.

\Needspace{2\baselineskip}
Existing visual-token methods reduce this cost through similarity-based merging~\cite{tome2023,holitom2025,unicomp2026}, attention-based pruning~\cite{fastv2024,visionzip2025,zhang2025sparsevlm}, or medical-specific reduction~\cite{hulu2025,medpruner2026}. Despite their effectiveness, two limitations remain. First, common similarity criteria compare normalized feature directions and therefore discard feature magnitude, while attention-based criteria inherit the attention pattern of a specific visual--language interface (i.e., the visual-encoder and connector configuration supplying tokens to the language decoder). This motivates a feature relation that uses both magnitude and direction. Second, content selection and capacity control are often coupled. Methods based on prescribed retention ratios keep the retention fraction fixed across interfaces~\cite{fastv2024,visionzip2025}, whereas adaptive rules allow similarity, clustering, or cumulative attention to determine both token identity and token count~\cite{prumerge2025,medpruner2026}. The former cannot adapt resource use, while the latter can produce substantially different sequence lengths when the underlying score distribution changes.

Volumetric images provide structure for addressing these limitations. Unlike videos that may contain abrupt changes between shots, consecutive medical slices share an in-plane coordinate system and repeatedly encode the same anatomy along depth. This recurrent context coexists with localized variations such as lesions, vessels, and anatomical boundaries. Our analysis finds stronger feature affinity between adjacent than nonadjacent slices across the evaluated datasets and backbones (Fig.~\ref{fig:motivation}(a)). Feature-driven composition at 10\%--30\% nominal retention matches or exceeds the uncompressed reference in several settings (Fig.~\ref{fig:motivation}(b)). In contrast, attention concentration varies markedly across visual--language interfaces (Fig.~\ref{fig:motivation}(c)--(f)). These observations indicate that volumetric compression should use cross-slice feature relations to determine \emph{what} content is represented, while controlling \emph{how many} tokens are retained through a separate interface-aware mechanism.

To this end, we propose \method{}, a training-free framework comprising two decoupled modules: hyperbolic token composition and attention-calibrated capacity control. The composition module determines \emph{what} content is represented. It organizes contiguous slices into ordered support units and uses the Lorentz distance to account for feature magnitude and direction during representative allocation, distinctive anchor selection, and count-preserving refinement. The capacity-control module independently determines \emph{how many} tokens are retained by providing either a prescribed budget or a bounded, interface-aware budget calibrated from visual-encoder attention entropy without an additional decoder pass. This separation lets attention set a global cap and feature relations determine the compact representation. 

The main contributions are summarized as follows:
\begin{itemize}
    \item We propose \method{}, a training-free and model-agnostic framework that decouples volumetric content composition from interface-aware capacity control.

    \item We design hyperbolic token composition over ordered support units, characterize how feature magnitude modulates directional differences, and derive a bound on refinement displacement.

    \item We design an attention-calibrated capacity control module that sets a bounded, interface-aware token budget from visual-encoder attention entropy.

    \item We evaluate \method{} on two tasks, three datasets, and four VLM backbones, demonstrating effective and efficient compression under fixed and calibrated token budgets.
\end{itemize}

\section{Related Work}
\label{sec:related_work}

\subsection{Volumetric Medical VLMs and Token Compression}
\label{sec:rw_vlms}
Vision--language representation learning~\cite{radford2021clip} and visual instruction tuning~\cite{liu2023llava} provide foundations for medical VLMs, supported by medical imaging resources~\cite{deng2026project,qu2023abdomenatlas} and modality-specific alignment~\cite{ye2026unveiling}. Most medical VLMs, such as LLaVA-Med~\cite{li2023llavamed}, HuatuoGPT-Vision~\cite{chen2024huatuogptvision}, and Lingshu~\cite{xu2025lingshu}, are designed for 2D images and do not support full-volume inputs. Volumetric models either adopt dedicated 3D encoders~\cite{m3d2024,wu2025radfm} or, as in Hulu-Med~\cite{hulu2025} and MedGemma-1.5~\cite{medgemma2026}, encode volumes slice by slice. Slice-based encoding exposes heterogeneous visual--language interfaces across models, and the visual sequence grows with the number of slices. Several methods were proposed to compress volumetric tokens. Hulu-Med~\cite{hulu2025} combines in-plane pooling with embedding-difference pruning across adjacent slices. Photon~\cite{photon2026} learns instruction-conditioned retention thresholds. MedARC~\cite{medarc2026} guides merging with encoder attention, visual--text relevance, and an auxiliary visual foundation model. ORCA~\cite{orca2026} merges spatially connected regions under organ guidance. CARVE~\cite{carve2026} allocates a prescribed budget across depth windows and retrieves complementary cross-slice evidence. MedPruner~\cite{medpruner2026} combines inter-slice filtering with adaptive within-slice token selection. Volumetric token compression involves deciding how to represent content across slices and how many tokens to retain. \method{} uses cross-slice feature relations to compose tokens and sets the retention fraction separately using visual-encoder attention entropy. This calibration allows the token budget to adapt to differences in attention dispersion across inputs and backbones.

\subsection{Feature Relations for Token Merging}
\label{sec:rw_hyperbolic}
Similarity-based methods merge redundant tokens. ToMe~\cite{tome2023} merges similar tokens within vision Transformers (ViTs), and video compressors merge across frames~\cite{fu2025framefusion,holitom2025}, including frame-group fusion with uniqueness-driven compression~\cite{unicomp2026}. Their similarity criteria are typically angular and thus blind to feature magnitude. Hyperbolic geometry provides relations in which magnitude matters. Hyperbolic space embeds hierarchies with low distortion because its volume grows exponentially with radius~\cite{lorentz2018}. It has been used to learn hierarchical structure in 3D biomedical images~\cite{hsu2021hierarchical} and image--text representations~\cite{desai2023meru}, and to merge tokens within vision Transformers~\cite{chen2026hyptm}. These approaches learn or operate on hyperbolic embeddings inside the model. \method{} instead applies the Lorentz relation, without training, to the native features passed from a frozen encoder to a frozen decoder. The Lorentz distance therefore guides the composition of compact volumetric representations.

\subsection{Attention-Based Selection and Budget Setting}
\label{sec:rw_token_efficiency}
Saliency-based methods rank tokens by attention. FastV~\cite{fastv2024} prunes with decoder attention, and SparseVLM~\cite{zhang2025sparsevlm} uses text-guided attention. VisionZip~\cite{visionzip2025} retains attention-dominant tokens and merges contextual ones, and HiPrune~\cite{hiprune2026} exploits hierarchical encoder attention. These criteria inherit the attention pattern of each interface, which varies markedly across medical VLMs (Fig.~\ref{fig:motivation}). Budgets, in turn, are either prescribed as fixed retention ratios or determined by the selection rule itself~\cite{prumerge2025,medpruner2026}. A common fixed ratio does not adapt to interface differences, whereas a selection-rule-determined budget couples token ranking with retention size. \method{} assigns each signal its own role. A magnitude-aware relation over native visual features decides which content is represented, and attention entropy sets the global token budget.

\Needspace{4\baselineskip}
\section{Method}
\label{sec:method}
\subsection{Problem Formulation and Framework Overview}
\label{sec:formulation}
\begin{figure*}[!t]\centering\includegraphics[width=0.97\textwidth]{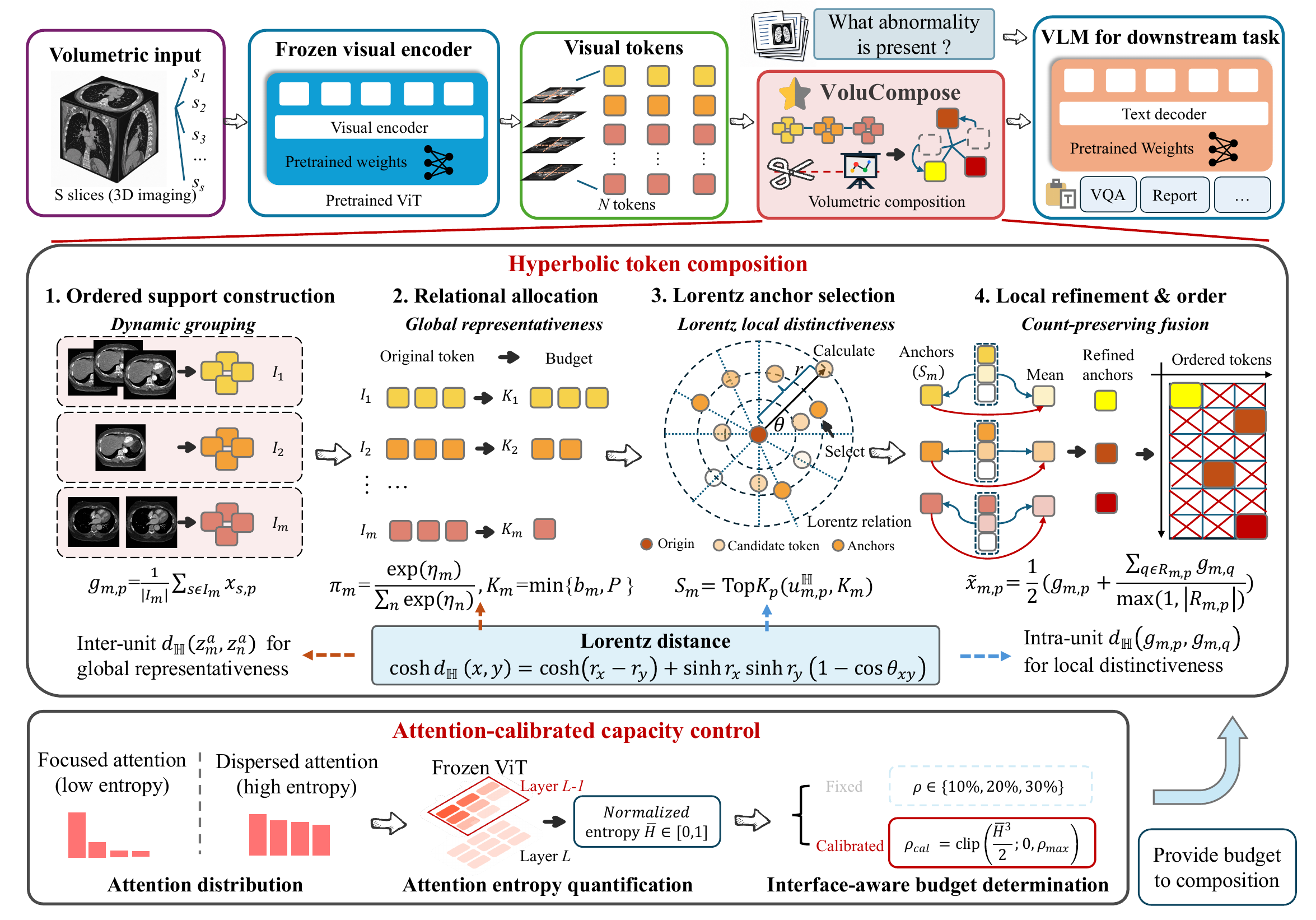}\caption{Overview of \method{}. A frozen visual interface converts an ordered volume into visual tokens. The hyperbolic token composition module groups contiguous slices into ordered support units, allocates tokens across units, selects distinctive anchors with the Lorentz distance, and refines them with related context while preserving source order. The attention-calibrated capacity control module determines the total budget $B$ from a prescribed retention fraction or normalized attention entropy. The compact token sequence is supplied to the language decoder for downstream tasks.}\label{fig:method_overview}\end{figure*}

For a volume of $S$ preprocessed slices, the visual encoder produces feature grids $F\in\mathbb R^{S\times H_{\mathrm p}\times W_{\mathrm p}\times d}$, where $H_{\mathrm p}$ and $W_{\mathrm p}$ are the patch-grid dimensions and $d$ is the feature dimension. Flattening the two spatial axes of slice $s$ gives $X_s=[x_{s,1},\ldots,x_{s,P}]^{\mathsf T}\in\mathbb R^{P\times d}$, with $P=H_{\mathrm p}W_{\mathrm p}$. The full visual sequence is $\mathcal X=[X_1;\ldots;X_S]\in\mathbb R^{N\times d}$, where $N=SP$. For a retention fraction $\rho\in[0,1]$, we set the global token budget to $B=\lfloor\rho N\rfloor$. A compression operator $\mathcal C$ then produces
\begin{equation}
\widetilde{\mathcal X}=\mathcal C(\mathcal X;B)\in\mathbb R^{L\times d},
\qquad L\leq B.
\label{eq:problem}
\end{equation}
As shown in Fig.~\ref{fig:method_overview}, \method{} comprises two decoupled modules. Hyperbolic token composition constructs $\widetilde{\mathcal X}$ from native visual features through ordered support construction, inter-unit token allocation, and anchor selection with count-preserving refinement. Attention-calibrated capacity control independently supplies the global budget $B$ from either a prescribed retention fraction or visual-encoder attention entropy.

\subsection{Hyperbolic Token Composition}
\label{sec:volumetric_composition}
\subsubsection{Preliminaries: Lorentz Geometry}
To compare features using both magnitude and direction, we use the Lorentz model of hyperbolic geometry~\cite{lorentz2018}. This model represents hyperbolic space as the upper sheet of a unit hyperboloid, $\mathbb H^h=\{u\in\mathbb R^{h+1}:\langle u,u\rangle_{\mathcal L}=-1,\ u_0>0\}$. For $x\in\mathbb R^h$, the exponential map at the origin is
\begin{equation}
\phi(x)=\left(\cosh r_x,\frac{\sinh r_x}{r_x}x\right),
\qquad r_x=\|x\|_2,
\label{eq:expmap}
\end{equation}
with $\phi(0)=(1,0,\ldots,0)$ by continuity. For mapped points, the Lorentz inner product is $\langle u,v\rangle_{\mathcal L}=-u_0v_0+\sum_{j=1}^{h}u_jv_j$, giving
\begin{equation}
d_{\mathbb H}(x,y)=\operatorname{arcosh}\bigl(-\langle\phi(x),\phi(y)\rangle_{\mathcal L}\bigr).
\label{eq:lorentz_distance}
\end{equation}
For nonzero features with $\cos\theta_{xy}=x^{\mathsf T}y/(r_xr_y)$,
\begin{equation}
\cosh d_{\mathbb H}(x,y)=\cosh(r_x-r_y)+\sinh r_x\sinh r_y(1-\cos\theta_{xy}).
\label{eq:lorentz_relation}
\end{equation}
This relation guides support construction, inter-unit token allocation, and anchor selection and refinement. Learned hyperbolic embeddings can associate radius with hierarchical specificity~\cite{lorentz2018,desai2023meru}. Here, radius supplies a magnitude cue from frozen native features. Composition itself operates in the native feature space consumed by the decoder. The relation thus determines which content is kept without altering the input space of the frozen decoder.

\subsubsection{Ordered Support Construction}
\label{sec:budget_formation}
Through-plane recurrence motivates aggregating redundant content across contiguous slices. We compute a slice descriptor $z_s^{\mathrm f}=P^{-1}\sum_p x_{s,p}\in\mathbb R^d$. Starting from the first unassigned slice $\ell_m$, we append consecutive slices while $d_{\mathbb H}(z_s^{\mathrm f},z_{\ell_m}^{\mathrm f})\leq\epsilon_{\mathrm f}$, where $\epsilon_{\mathrm f}\geq0$ is the grouping threshold. Repeating this procedure partitions the sequence into $M$ ordered support units $\mathcal I_m$. Position-wise averaging~\cite{unicomp2026} yields
\begin{equation}
g_{m,p}=\frac{1}{|\mathcal I_m|}\sum_{s\in\mathcal I_m}x_{s,p},
\qquad G_m=[g_{m,1},\ldots,g_{m,P}]^{\mathsf T}.
\label{eq:support_aggregation}
\end{equation}
Because contiguous slices share an in-plane coordinate frame, each grid position follows the same anatomical column through the unit. The radius of the aggregated feature therefore records cross-slice consistency. By the triangle inequality,
\begin{equation}
\|g_{m,p}\|_2\leq\frac{1}{|\mathcal I_m|}\sum_{s\in\mathcal I_m}\|x_{s,p}\|_2,
\label{eq:consistency_radius}
\end{equation}
with equality when all nonzero contributing features are positively collinear. Directional disagreement reduces the radius relative to the average slice-wise magnitude. The Lorentz distance retains this magnitude information, whereas cosine similarity discards it.

\subsubsection{Inter-Unit Token Allocation}
Given the global budget $B$, we allocate tokens across support units using their feature relations. We pool channels into a patch profile $z_m^{\mathrm a}\in\mathbb R^P$, with $z_m^{\mathrm a}[p]=d^{-1}\sum_j g_{m,p,j}$. The mean relation distances and allocation logits are
\begin{equation}
\begin{aligned}
\bar A_m&=\frac{1}{M}\sum_{n=1}^{M}d_{\mathbb H}(z_m^{\mathrm a},z_n^{\mathrm a}),
\qquad \bar A=\frac{1}{M}\sum_m\bar A_m,\\
\eta_m&=\sqrt M(\bar A-\bar A_m),
\qquad \pi_m=\frac{e^{\eta_m}}{\sum_n e^{\eta_n}}.
\end{aligned}
\label{eq:allocation_logit}
\end{equation}
To limit budget concentration in individual support units, we impose the per-unit capacity cap $c=\min\{C_{\max},\max(1,2\lfloor B/M\rfloor)\}$, where $C_{\max}$ is a configured positive integer limit. If $Mc\leq B$, we set $b_m=c$; otherwise, we initialize $b_m^{(0)}=\min\{c,\lfloor B\pi_m\rfloor\}$. When clipping occurs, residual tokens are redistributed in equal integer batches among eligible units. Redistribution stops when no new clipping occurs or no full batch can be assigned. The resulting capacities satisfy $\sum_m b_m\leq B$, and each unit retains $K_m=\min\{b_m,P\}$ anchors.

Allocation and anchor selection follow complementary principles. Across support units, smaller average relation distances yield larger allocation weights. Allocation therefore favors representative anatomical context and limits the capacity spent on atypical, repetitive units, such as those at the start of a volume. Within each unit, lower mean normalized similarity yields higher distinctiveness, allowing localized feature variation to compete for its capacity.

\subsubsection{Anchor Selection and Count-Preserving Refinement}
\label{sec:semantic_depth}
Within each unit, define $D_{m,pq}=d_{\mathbb H}(g_{m,p},g_{m,q})$. Let $D^{\max}_{m,p}$ and $D^{\min}_{m,p}$ denote its row extrema over all patches, including the diagonal. We normalize negative distances and score token distinctiveness by
\begin{equation}
\begin{aligned}
S^{\mathbb H}_{m,pq}&=\frac{D^{\max}_{m,p}-D_{m,pq}}
{D^{\max}_{m,p}-D^{\min}_{m,p}+\varepsilon_{\mathrm{rel}}},\\
u^{\mathbb H}_{m,p}&=1-\frac{1}{P}\sum_{q=1}^{P}S^{\mathbb H}_{m,pq},
\end{aligned}
\label{eq:relation_normalization}
\end{equation}
where $\varepsilon_{\mathrm{rel}}=10^{-8}$ is a small constant for numerical stability. A smaller mean normalized similarity gives a higher distinctiveness score. The anchor set $\mathcal S_m=\operatorname{TopK}_{p}(u^{\mathbb H}_{m,p},K_m)$ contains the $K_m$ highest-scoring indices. The retained count is thus fixed before refinement, without greedy neighbor exclusion.

For each selected anchor, a relation threshold $\tau_{\mathrm{nbr}}\in(0,1]$ defines the neighborhood $\mathcal R_{m,p}=\{q:1-S^{\mathbb H}_{m,pq}<\tau_{\mathrm{nbr}}\}$. Neighborhoods are defined in feature-relation space and may overlap across anchors. Refinement combines each anchor with its related context while preserving the selected token count:
\begin{equation}
\widetilde x_{m,p}=\frac{1}{2}\left(g_{m,p}+
\frac{\sum_{q\in\mathcal R_{m,p}}g_{m,q}}
{\max(1,|\mathcal R_{m,p}|)}\right),
\qquad p\in\mathcal S_m.
\label{eq:neighbor_fusion}
\end{equation}
Refined anchors are concatenated in support-unit and source-patch order, and their count remains $L=\sum_m K_m$.

\subsection{Attention-Calibrated Capacity Control}
\label{sec:capacity}
The attention-calibrated capacity control module determines the global token budget $B$ independently of content composition. It accepts a prescribed retention fraction or calibrates a bounded fraction from each backbone's visual-encoder attention concentration. For calibration, we use self-attention from the penultimate layer of the frozen visual encoder. These maps are produced during visual encoding, so calibration requires neither a decoder pass nor the question text. For $R$ nonnegative attention rows $\alpha^{(t)}$ with positive sums and lengths $N_t>1$, define $a_i^{(t)}=\alpha_i^{(t)}/\sum_j\alpha_j^{(t)}$ and the mean normalized entropy
\begin{equation}
\bar H=-\frac{1}{R}\sum_{t=1}^{R}
\frac{\sum_{i=1}^{N_t}a_i^{(t)}\log(a_i^{(t)}+\varepsilon_{\mathrm H})}{\log N_t},
\label{eq:attention_entropy}
\end{equation}
where $\varepsilon_{\mathrm H}=10^{-12}$ is a small constant for numerical stability. Ignoring this constant, $N_t^{H_t}$ gives the effective number of attended tokens for a row with normalized entropy $H_t$. Thus, $\bar H$ measures attention dispersion rather than token identity. A bounded cubic rule supplies
\begin{equation}
\rho_{\mathrm{cal}}=\operatorname{clip}_{[0,\rho_{\max}]}\left(\frac{\bar H^3}{2}\right),
\qquad B=\lfloor\rho_{\mathrm{cal}}N\rfloor,
\label{eq:car}
\end{equation}
where $\rho_{\max}$ caps the retention fraction. On $[0,1]$, the uncapped cubic is monotone and convex; clipping preserves monotonicity and enforces the cap. Concentrated attention receives smaller budgets, while dispersed attention receives larger budgets up to the cap. The factor one half limits uncapped retention to 50\% under uniform attention. The budget $B$ is determined by attention, while native features determine support construction, allocation, and anchor scores. Selection fixes $L=\sum_mK_m\leq B$ before refinement, independently of $\tau_{\mathrm{nbr}}$. For the same input, composition parameters, and $B$, both paths yield identical representations.

\subsection{Theoretical Properties of Hyperbolic Token Composition}
\label{sec:unified}
\textit{Properties of the Lorentz distance.} Let $s_{xy}=\sin(\theta_{xy}/2)$, so that $1-\cos\theta_{xy}=2s_{xy}^2$. The cosine, Euclidean, and Lorentz relations then share a common form:
\begin{equation}
\begin{aligned}
1-\cos\theta_{xy}&=2s_{xy}^2,\\
\|x-y\|_2^2&=(r_x-r_y)^2+4r_xr_y\,s_{xy}^2,\\
\cosh d_{\mathbb H}(x,y)&=\cosh(r_x-r_y)+2\sinh r_x\sinh r_y\,s_{xy}^2.
\end{aligned}
\label{eq:coupling}
\end{equation}
All three compare feature directions. They differ, however, in how feature magnitude gates angular differences. Cosine similarity applies no gating, the Euclidean distance applies bilinear gating, and the Lorentz relation applies exponential gating in $\cosh d_{\mathbb H}$. Two regimes follow. As $r_x,r_y\to0$ with $x\neq y$, $d_{\mathbb H}(x,y)/\|x-y\|_2\to1$, so small-magnitude features are compared approximately in Euclidean terms. For $r_x,r_y\gg1$ and $2\sinh r_x\sinh r_y\,s_{xy}^2\gg\cosh(r_x-r_y)$,
\begin{equation}
d_{\mathbb H}(x,y)\approx r_x+r_y+2\log s_{xy},
\label{eq:large_radius}
\end{equation}
so radii contribute additively and angular separation contributes logarithmically. In this regime, large-radius features remain separated unless their directions are nearly collinear. Because the Euclidean distance is also sensitive to magnitude, the distinguishing property of the Lorentz relation is the exponential gating of angular differences by magnitude. Moreover,
\begin{equation}
d_{\mathbb H}(x,y)\geq\|x-y\|_2\geq|r_x-r_y|.
\label{eq:metric_control}
\end{equation}
For a short proof, let $w=(1-\cos\theta_{xy})/2$. Then $\|x-y\|_2^2=(1-w)(r_x-r_y)^2+w(r_x+r_y)^2$. Convexity of $z\mapsto\cosh\sqrt z$ bounds $\cosh\|x-y\|_2$ by $(1-w)\cosh(r_x-r_y)+w\cosh(r_x+r_y)=\cosh d_{\mathbb H}(x,y)$. Monotonicity of $\cosh$ then proves the first inequality. The second is the reverse triangle inequality, and zero features follow by continuity.

\textit{Anchor fidelity and controlled refinement.} For a nonempty neighborhood, write $g=g_{m,p}$, $\mathcal R=\mathcal R_{m,p}$, and $\mu=|\mathcal R|^{-1}\sum_{q\in\mathcal R}g_{m,q}$. The update in Eq.~\eqref{eq:neighbor_fusion} uniquely minimizes
\begin{equation}
Q(v)=\|v-g\|_2^2+
\frac{1}{|\mathcal R|}\sum_{q\in\mathcal R}\|v-g_{m,q}\|_2^2.
\label{eq:refinement_objective}
\end{equation}
For $v\in\mathbb R^d$, $\nabla Q(v)=4v-2(g+\mu)$ and $\nabla^2Q(v)=4I_d$, where $I_d$ is the identity matrix. Thus, the minimizer $(g+\mu)/2$ balances anchor fidelity and neighborhood fit. Since $D^{\min}_{m,p}=0$, define $\delta_{m,p}=\tau_{\mathrm{nbr}}(D^{\max}_{m,p}+\varepsilon_{\mathrm{rel}})-\varepsilon_{\mathrm{rel}}$. Membership in $\mathcal R$ implies $D_{m,pq}<\delta_{m,p}$. Equation~\eqref{eq:metric_control} and the triangle inequality therefore give
\begin{equation}
\|\widetilde x_{m,p}-g\|_2
\leq\frac{\sum_{q\in\mathcal R}D_{m,pq}}{2|\mathcal R|}
<\frac{\delta_{m,p}}{2}.
\label{eq:fusion_bound}
\end{equation}
For nonempty neighborhoods, this bound controls displacement from the aggregated anchor within the native feature space. Compatibility with a frozen decoder is assessed empirically in Section~\ref{sec:components}.

\subsection{Implementation and Computational Characteristics}
\label{sec:implementation}
The reference implementation requires $O(SPd+S^2d+M^2P+MP^2d)$ leading arithmetic operations for pooling, slice relations, unit allocation, and token composition. It stores $O(S^2+M^2+MP^2)$ pairwise values in addition to feature tensors. Neither path uses the question text, so a compressed study can be reused across multiple questions. Decoder-prefill floating-point operations (FLOPs) and key--value (KV) cache demand are analyzed in Section~\ref{sec:cost}.

\section{Experiments}
\label{sec:results}
\begin{table*}[!t]
\caption{Performance at 10\% nominal retention. Under the fixed protocol, dataset performance is averaged over four backbones, while overall ROUGE-L and closed-question accuracy (Close-Acc) are averaged across 12 and eight settings, respectively. Worst case is the lowest setting-level score as a percentage of the corresponding uncompressed score. Bold marks the best compression result in each column.}
\label{tab:fixedoverview}\centering\tableformat
\begin{tabularx}{\textwidth}{@{}l*{10}{>{\centering\arraybackslash}X}@{}}
\toprule
& \multicolumn{2}{c}{3D-RAD~\cite{rad2025}} & \multicolumn{2}{c}{M3D~\cite{m3d2024}} & \multicolumn{2}{c}{AMOS-MM~\cite{amos2024}} & \multicolumn{2}{c}{Overall} & \multicolumn{2}{c}{Worst case (\%)} \\
\cmidrule(lr){2-3}\cmidrule(lr){4-5}\cmidrule(lr){6-7}\cmidrule(lr){8-9}\cmidrule(l){10-11}
Method & ROUGE-L & Close-Acc & ROUGE-L & Close-Acc & ROUGE-L & METEOR & ROUGE-L & Close-Acc & ROUGE-L & Close-Acc \\
\midrule
Uncompressed & 18.09 & 55.62 & 25.34 & 55.83 & 19.01 & 18.91 & 20.81 & 55.73 & -- & -- \\
\midrule
HoliTom~\cite{holitom2025} & 21.50 & 52.57 & \textbf{29.31} & 57.96 & 17.57 & 13.28 & 22.79 & 55.27 & 37.7 & 84.0 \\
VisionZip~\cite{visionzip2025} & 21.43 & 53.93 & 28.99 & \textbf{59.04} & 15.91 & 11.83 & 22.11 & 56.48 & 15.0 & 81.3 \\
UniComp~\cite{unicomp2026} & 21.15 & 51.31 & 25.39 & 55.54 & 17.63 & 14.36 & 21.39 & 53.42 & 70.5 & 78.8 \\
\rowcolor{PaperHighlight}VoluCompose & \textbf{21.67} & \textbf{56.65} & 28.89 & 57.78 & \textbf{23.57} & \textbf{17.91} & \textbf{24.71} & \textbf{57.22} & \textbf{96.4} & \textbf{94.2} \\
\bottomrule
\end{tabularx}
\end{table*}

\begin{table*}[!t]
\caption{Per-backbone performance at 10\% nominal retention. Visual question answering (VQA) is evaluated on 3D-RAD and M3D, and report generation on AMOS-MM. Bold marks the highest score among compression methods in each column and backbone.}
\label{tab:fixeddetail}\centering\tableformat\renewcommand{\arraystretch}{1.02}
\begin{tabularx}{\textwidth}{@{}l*{9}{>{\centering\arraybackslash}X}@{}}
\toprule
& \multicolumn{2}{c}{3D-RAD~\cite{rad2025}} & \multicolumn{2}{c}{M3D~\cite{m3d2024}} & \multicolumn{5}{c}{AMOS-MM~\cite{amos2024}} \\
\cmidrule(lr){2-3}\cmidrule(lr){4-5}\cmidrule(l){6-10}
Method & ROUGE-L & Close-Acc & ROUGE-L & Close-Acc & ROUGE-L & METEOR & ROUGE-1 & BLEU-1 & BLEU-4 \\
\midrule
\multicolumn{10}{@{}l}{\textit{Hulu-7B}} \\
Uncompressed & 22.92 & 78.77 & 34.05 & 75.32 & 28.76 & 31.73 & 39.52 & 37.32 & 13.38 \\
\addlinespace[0.5pt]
HoliTom~\cite{holitom2025} & \textbf{31.86} & 77.94 & 46.15 & 76.46 & 27.21 & 19.52 & 28.56 & 23.16 & 5.89 \\
VisionZip~\cite{visionzip2025} & 31.47 & \textbf{78.84} & 45.19 & 75.65 & 26.60 & 18.48 & 27.99 & 22.53 & 4.95 \\
UniComp~\cite{unicomp2026} & 30.36 & 76.23 & 41.46 & 74.14 & 20.28 & 13.69 & 21.38 & 14.43 & 2.44 \\
\rowcolor{PaperHighlight}VoluCompose & 31.11 & 78.70 & \textbf{46.81} & \textbf{77.02} & \textbf{37.41} & \textbf{28.35} & \textbf{39.42} & \textbf{34.60} & \textbf{11.52} \\
\midrule
\multicolumn{10}{@{}l}{\textit{MedGemma-1.5-4B}} \\
Uncompressed & 4.09 & 59.16 & 5.89 & 32.72 & 9.91 & 9.72 & 13.79 & 5.85 & 0.25 \\
\addlinespace[0.5pt]
HoliTom~\cite{holitom2025} & 8.66 & 57.58 & 8.54 & 40.92 & 16.53 & 9.37 & 17.57 & 8.40 & 0.77 \\
VisionZip~\cite{visionzip2025} & 8.23 & 55.66 & \textbf{9.18} & \textbf{45.41} & 15.02 & 7.57 & 15.83 & 4.77 & 0.31 \\
UniComp~\cite{unicomp2026} & 8.22 & \textbf{59.47} & 8.79 & 39.15 & 17.46 & \textbf{13.08} & 18.28 & \textbf{13.26} & \textbf{1.20} \\
\rowcolor{PaperHighlight}VoluCompose & \textbf{8.89} & 57.80 & 8.97 & 39.72 & \textbf{17.99} & 10.56 & \textbf{19.18} & 10.38 & 0.84 \\
\midrule
\multicolumn{10}{@{}l}{\textit{Qwen2.5-VL-7B}} \\
Uncompressed & 22.05 & 34.17 & 27.96 & 58.33 & 19.04 & 15.37 & 20.76 & 20.38 & 0.99 \\
\addlinespace[0.5pt]
HoliTom~\cite{holitom2025} & 22.28 & 28.70 & \textbf{28.87} & 57.80 & 7.17 & 6.31 & 7.66 & 8.85 & 0.58 \\
VisionZip~\cite{visionzip2025} & 22.80 & 27.79 & 28.62 & \textbf{58.10} & 2.85 & 2.65 & 3.12 & 3.47 & 0.16 \\
UniComp~\cite{unicomp2026} & 22.27 & 29.80 & 21.58 & 56.08 & 15.90 & 13.71 & 16.94 & 16.33 & 0.49 \\
\rowcolor{PaperHighlight}VoluCompose & \textbf{23.11} & \textbf{42.61} & 26.95 & 57.79 & \textbf{19.64} & \textbf{15.46} & \textbf{21.26} & \textbf{20.77} & \textbf{1.51} \\
\midrule
\multicolumn{10}{@{}l}{\textit{Qwen3-VL-8B}} \\
Uncompressed & 23.31 & 50.39 & 33.45 & 56.96 & 18.32 & 18.82 & 18.72 & 26.99 & 1.08 \\
\addlinespace[0.5pt]
HoliTom~\cite{holitom2025} & 23.19 & 46.04 & \textbf{33.66} & 56.68 & \textbf{19.38} & 17.91 & \textbf{21.23} & 25.87 & \textbf{1.37} \\
VisionZip~\cite{visionzip2025} & 23.20 & \textbf{53.43} & 32.98 & \textbf{56.99} & 19.16 & \textbf{18.62} & 20.92 & \textbf{27.18} & 1.16 \\
UniComp~\cite{unicomp2026} & \textbf{23.73} & 39.73 & 29.73 & 52.79 & 16.88 & 16.97 & 18.51 & 23.49 & 0.68 \\
\rowcolor{PaperHighlight}VoluCompose & 23.55 & 47.49 & 32.84 & 56.60 & 19.25 & 17.27 & 21.14 & 24.99 & 1.21 \\
\bottomrule
\end{tabularx}
\end{table*}

\subsection{Experimental Setup}
\label{sec:setup}
We evaluate \method{} on M3D~\cite{m3d2024} and 3D-RAD~\cite{rad2025} for visual question answering (VQA) and on AMOS-MM~\cite{amos2024} for report generation on their evaluation splits. The four backbones are Hulu-7B~\cite{hulu2025}, MedGemma-1.5-4B~\cite{medgemma2026}, Qwen2.5-VL-7B~\cite{qwen252025}, and Qwen3-VL-8B~\cite{qwen32025}. We use dataset/backbone notation, such as AMOS-MM/Hulu-7B, throughout.

The decoupled design allows the same volumetric composition procedure to operate under prescribed or attention-calibrated budgets. We therefore evaluate compression under two complementary protocols. The fixed protocol compares methods at common nominal retention fractions. The calibrated protocol compares performance and realized retention at the reported operating points.

Under the fixed protocol, \method{} is compared with HoliTom~\cite{holitom2025}, VisionZip~\cite{visionzip2025}, and UniComp~\cite{unicomp2026} at nominal retention fractions of 10\%, 20\%, and 30\% across 12 settings. Under the calibrated protocol, \method{} uses $\rho_{\max}=0.30$. This cap sets an explicit limit on visual-token retention, equal to the largest evaluated fixed fraction. Attention determines the budget within this limit, and calibrated fractions can fall below 10\%. \method{} is compared with Hulu-Med L1~\cite{hulu2025}, HiPrune~\cite{hiprune2026}, and MedPruner~\cite{medpruner2026} across nine settings on Hulu-7B, MedGemma-1.5-4B, and Qwen3-VL-8B. HiPrune~\cite{hiprune2026} is included in the calibrated comparison because its token budget scales with the native visual-token count. Its transferred setting yields about 22.30\% retention in most settings.

For VQA, we report closed-question accuracy (Close-Acc) and the longest common subsequence variant of Recall-Oriented Understudy for Gisting Evaluation (ROUGE-L)~\cite{rouge2004}. Report generation is evaluated with Bilingual Evaluation Understudy (BLEU) scores at orders one and four (BLEU-1 and BLEU-4)~\cite{bleu2002}, the unigram ROUGE score (ROUGE-1) and ROUGE-L~\cite{rouge2004}, and Metric for Evaluation of Translation with Explicit ORdering (METEOR)~\cite{meteor2005}. Text-generation metrics are reported on a 0--100 scale, and Close-Acc is a percentage. \textit{Uncompressed} denotes the reference scores for each backbone without visual-token compression. We exclude these references from method rankings.

Relative quality $Q=(\prod_{k=1}^{K}m_k/m_k^{(0)})^{1/K}$ is the geometric mean of metric ratios to the uncompressed scores $m_k^{(0)}$ in the same setting. We include both VQA metrics ($K=2$) and all five report metrics ($K=5$). At realized retention $r$, the single-point hypervolume (HV) in retention--quality space with reference $(1,0)$ is $(1-r)Q$. We compute $Q$ and HV per setting before averaging.

\begin{table*}[!t]
\caption{Performance and realized token retention at calibrated operating points. $Q$ and HV use the corresponding uncompressed scores. Overall columns average the settings in each block. Bold marks the highest $Q$ and HV among compressors.}
\label{tab:adaptivedetail}\centering\tableformat
\begin{tabularx}{\textwidth}{@{}l*{11}{>{\centering\arraybackslash}X}@{}}
\toprule
& \multicolumn{3}{c}{3D-RAD~\cite{rad2025}} & \multicolumn{3}{c}{M3D~\cite{m3d2024}} & \multicolumn{3}{c}{AMOS-MM~\cite{amos2024}} & \multicolumn{2}{c}{Overall} \\
\cmidrule(lr){2-4}\cmidrule(lr){5-7}\cmidrule(lr){8-10}\cmidrule(l){11-12}
Method & \shortstack{Retention\\rate (\%)} & ROUGE-L & Close-Acc & \shortstack{Retention\\rate (\%)} & ROUGE-L & Close-Acc & \shortstack{Retention\\rate (\%)} & ROUGE-L & METEOR & $Q$ (\%) & HV \\
\midrule
\multicolumn{12}{@{}l}{\textit{Hulu-7B}} \\
Uncompressed & 100.00 & 22.92 & 78.77 & 100.00 & 34.05 & 75.32 & 100.00 & 28.76 & 31.73 & 100.00 & 0.000 \\
\addlinespace[0.5pt]
Hulu-Med L1~\cite{hulu2025} & 45.06 & 23.49 & 78.08 & 66.03 & 47.17 & 77.79 & 16.35 & 24.83 & 26.21 & 100.65 & 0.547 \\
HiPrune~\cite{hiprune2026} & 22.30 & 25.56 & 79.06 & 22.30 & 46.20 & 77.62 & 22.30 & 27.96 & 31.09 & 106.91 & \textbf{0.831} \\
MedPruner~\cite{medpruner2026} & 51.88 & 25.98 & 79.28 & 52.10 & 47.26 & 77.45 & 54.20 & 28.86 & 31.14 & 108.50 & 0.514 \\
\rowcolor{PaperHighlight}VoluCompose & 29.46 & 31.41 & 78.26 & 29.69 & 47.60 & 77.47 & 29.94 & 39.42 & 29.59 & \textbf{113.03} & 0.795 \\
\midrule
\multicolumn{12}{@{}l}{\textit{MedGemma-1.5-4B}} \\
Uncompressed & 100.00 & 4.09 & 59.16 & 100.00 & 5.89 & 32.72 & 100.00 & 9.91 & 9.72 & 100.00 & 0.000 \\
\addlinespace[0.5pt]
Hulu-Med L1~\cite{hulu2025} & 44.89 & 1.60 & 57.87 & 65.98 & 4.47 & 44.64 & 16.07 & 9.79 & 9.78 & 87.59 & 0.506 \\
HiPrune~\cite{hiprune2026} & 22.30 & 2.05 & 55.96 & 22.30 & 4.45 & 44.42 & 21.85 & 9.78 & 9.94 & 89.50 & 0.697 \\
MedPruner~\cite{medpruner2026} & 4.62 & 2.06 & 60.84 & 4.87 & 5.85 & 43.72 & 2.46 & 10.23 & 10.25 & 115.76 & 1.115 \\
\rowcolor{PaperHighlight}VoluCompose & 9.57 & 8.31 & 58.80 & 8.78 & 9.12 & 38.39 & 8.59 & 17.92 & 10.54 & \textbf{151.84} & \textbf{1.383} \\
\midrule
\multicolumn{12}{@{}l}{\textit{Qwen3-VL-8B}} \\
Uncompressed & 100.00 & 23.31 & 50.39 & 100.00 & 33.45 & 56.96 & 100.00 & 18.32 & 18.82 & 100.00 & 0.000 \\
\addlinespace[0.5pt]
Hulu-Med L1~\cite{hulu2025} & 51.96 & 18.03 & 46.94 & 52.21 & 34.00 & 56.22 & 16.50 & 17.81 & 17.79 & 92.64 & 0.554 \\
HiPrune~\cite{hiprune2026} & 22.30 & 17.61 & 49.04 & 22.30 & 33.04 & 57.11 & 22.30 & 17.26 & 18.60 & 93.16 & \textbf{0.724} \\
MedPruner~\cite{medpruner2026} & 85.88 & 23.24 & 50.09 & 67.61 & 33.35 & 56.72 & 43.28 & 17.91 & 17.65 & 98.32 & 0.335 \\
\rowcolor{PaperHighlight}VoluCompose & 29.29 & 23.59 & 48.04 & 29.89 & 33.27 & 56.81 & 29.80 & 19.33 & 18.58 & \textbf{100.15} & 0.704 \\
\midrule
\multicolumn{12}{@{}l}{\textit{Mean across backbones}} \\
Hulu-Med L1~\cite{hulu2025} & 47.30 & 14.37 & 60.96 & 61.41 & 28.55 & 59.55 & 16.31 & 17.48 & 17.93 & 93.63 & 0.536 \\
HiPrune~\cite{hiprune2026} & 22.30 & 15.07 & 61.35 & 22.30 & 27.90 & 59.72 & 22.15 & 18.33 & 19.88 & 96.53 & 0.750 \\
MedPruner~\cite{medpruner2026} & 47.46 & 17.09 & 63.40 & 41.53 & 28.82 & 59.30 & 33.31 & 19.00 & 19.68 & 107.52 & 0.655 \\
\rowcolor{PaperHighlight}VoluCompose & 22.77 & 21.10 & 61.70 & 22.79 & 30.00 & 57.56 & 22.78 & 25.56 & 19.57 & \textbf{121.68} & \textbf{0.961} \\
\bottomrule
\end{tabularx}
\end{table*}

\begin{figure*}[!t]
\centering\includegraphics[width=.95\textwidth]{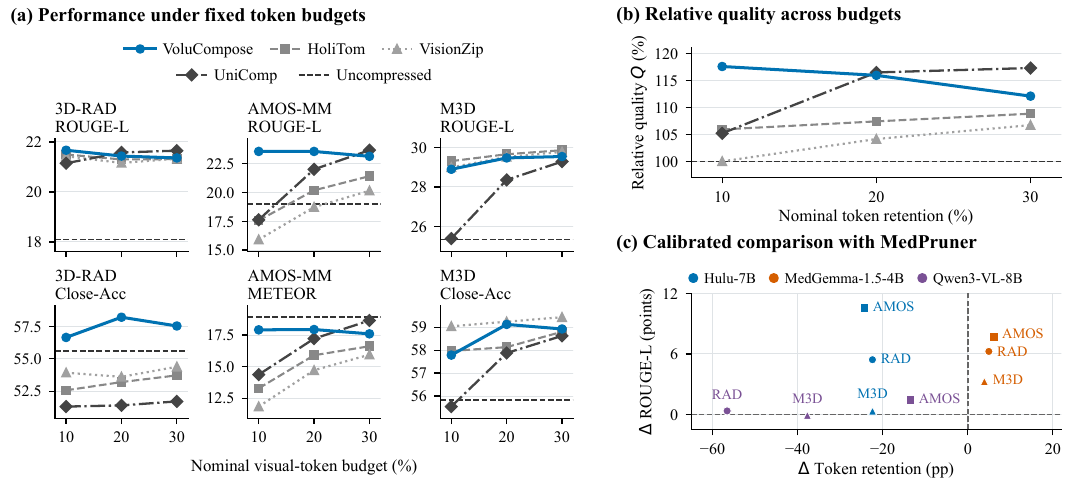}
\caption{Performance under fixed and calibrated token budgets. (a) Dataset means over four backbones. (b) Mean relative quality $Q$ across 12 fixed-budget settings per budget. Dashed lines in (a) and (b) mark uncompressed references. (c) Differences in ROUGE-L and realized token retention (\method{} minus MedPruner~\cite{medpruner2026}) across nine calibrated settings. Upper-left points have higher ROUGE-L and lower retention. pp: percentage points.}
\label{fig:budget_adaptive}
\end{figure*}

\begin{figure}[!t]
\centering\includegraphics[width=0.96\columnwidth]{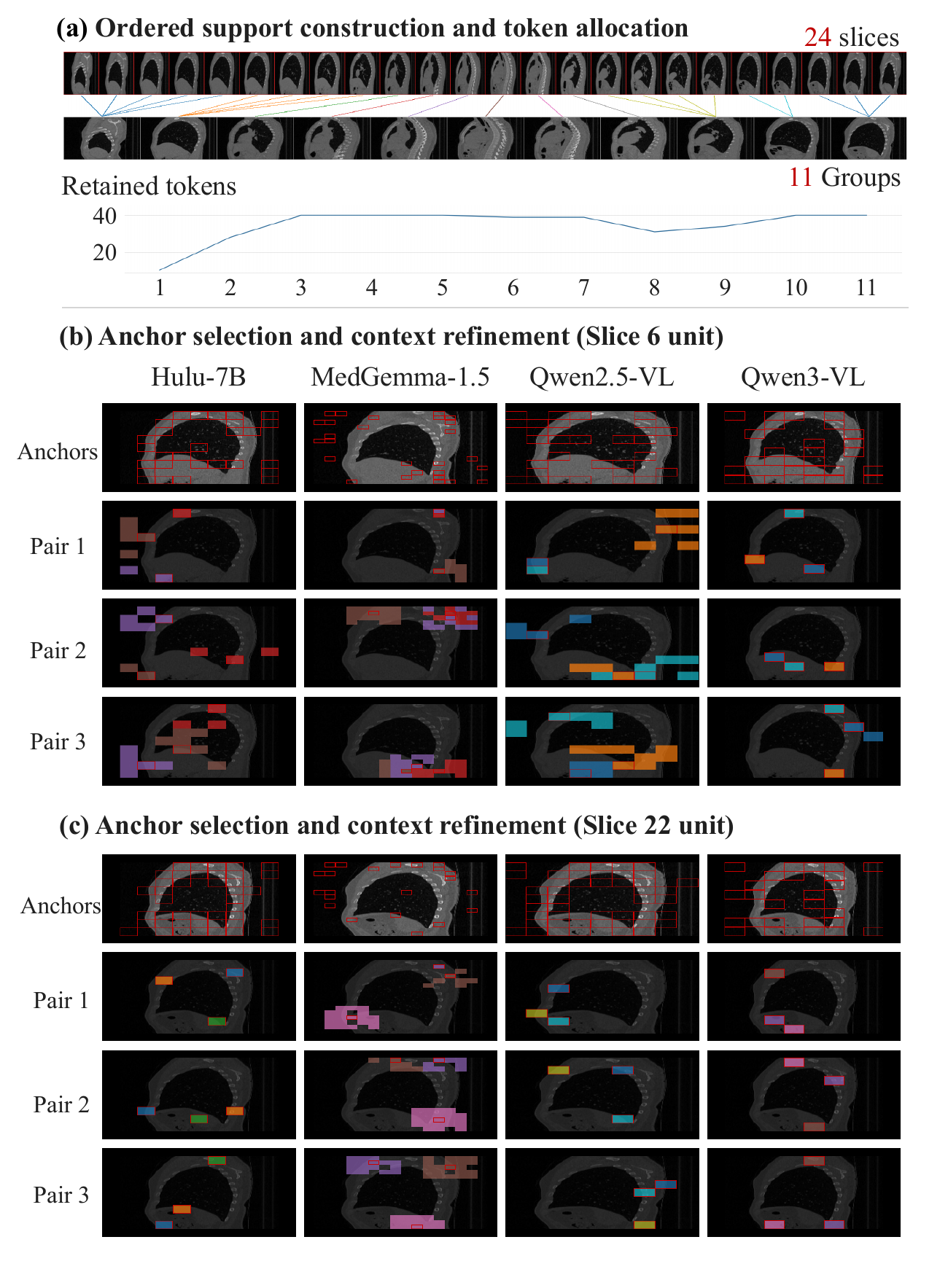}
\caption{Ordered support construction and anchor refinement. (a) A 24-slice Hulu-7B~\cite{hulu2025} input forms 11 ordered support units; the curve shows per-unit token allocation. (b), (c) Support units containing slices 6 and 22 across four backbones. The \textit{Anchors} row marks selected anchors in red. The \textit{Pair 1--3} rows show anchor--neighborhood examples in matching colors. Related patches can be spatially separated; refinement preserves the selected token count.}
\label{fig:process_example}
\end{figure}

\begin{figure}[!t]
\centering
\includegraphics[width=0.94\columnwidth]{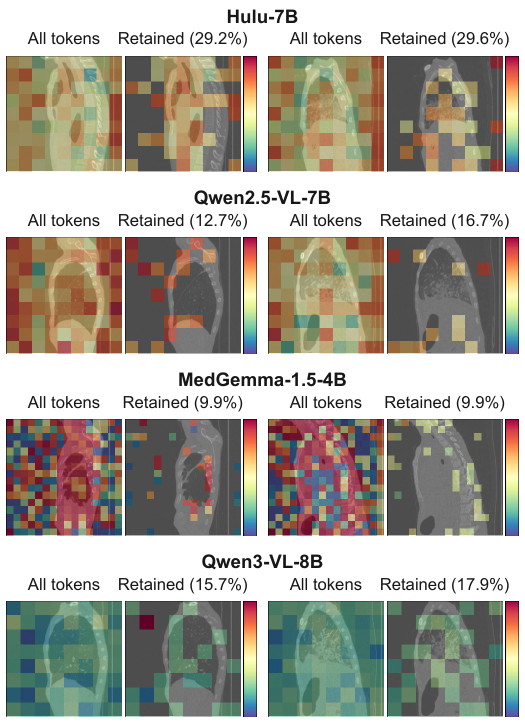}
\caption{Visual--text correlation maps for all and retained tokens. Each row shows two examples from one backbone. Paired maps use the same full text and color scale; gray marks excluded positions. Percentages indicate whole-volume token retention.}
\label{fig:semantic_examples}
\end{figure}

\subsection{Performance under Fixed Token Budgets}
\label{sec:fixed_results}
At 10\% nominal retention, \method{} achieved the highest mean ROUGE-L (24.71) and Close-Acc (57.22\%) among compressors (Tables~\ref{tab:fixedoverview} and~\ref{tab:fixeddetail}). It led both dataset-level metrics on 3D-RAD and AMOS-MM. ROUGE-L exceeded the uncompressed reference in 10 of 12 settings, including all 3D-RAD and AMOS-MM settings.

The lowest ratios of \method{} to the uncompressed references were 96.4\% for ROUGE-L and 94.2\% for Close-Acc (Table~\ref{tab:fixedoverview}). The baselines had minimum ROUGE-L ratios of 15.0\% to 70.5\%. The largest losses for HoliTom and VisionZip occurred with Qwen2.5-VL-7B, the backbone with the most dispersed attention profile (Fig.~\ref{fig:motivation}(e)). On AMOS-MM/Qwen2.5-VL-7B, their ROUGE-L fell to 37.7\% and 15.0\% of the uncompressed reference, respectively.

Across the three budgets, the largest overall ROUGE-L advantage occurred at 10\% (Fig.~\ref{fig:budget_adaptive}(a)). \method{} also led overall ROUGE-L and Close-Acc at 20\% and Close-Acc at 30\%. Its mean ROUGE-L stayed within 24.68--24.82, while the strongest baseline increased from 22.79 to 24.87. Relative quality $Q$ was highest for \method{} at 10\% and for UniComp at 20\% and 30\% (Fig.~\ref{fig:budget_adaptive}(b)).

The dataset pattern was consistent with through-plane recurrence: 3D-RAD and AMOS-MM showed stronger adjacent-slice affinity than M3D (Fig.~\ref{fig:motivation}(a)), and \method{} led their dataset-averaged summary metrics at 10\% retention. On M3D, it exceeded the uncompressed ROUGE-L reference for Hulu-7B and MedGemma-1.5-4B, while HoliTom led on the dataset mean. This association supports the motivation for ordered aggregation, although dataset and task differences preclude attributing it to recurrence alone.

\subsection{Attention-Calibrated Capacity}
\label{sec:dynamic_results}
At similar mean realized retention (22.78\% versus 22.25\%), \method{} reached higher mean relative quality (121.68\% versus 96.53\%) and HV (0.961 versus 0.750) than HiPrune (Table~\ref{tab:adaptivedetail}). \method{} led on mean HV, with the largest gain on MedGemma-1.5-4B. HiPrune led on Hulu-7B and Qwen3-VL-8B. MedPruner~\cite{medpruner2026} also adapts token retention for volumetric medical VLMs and had the highest mean $Q$ among the calibrated baselines. Fig.~\ref{fig:budget_adaptive}(c) shows a descriptive comparison at the reported operating points. \method{} had higher ROUGE-L in eight of nine settings, including five with lower retention. It ranked first in ROUGE-L among compression configurations in eight of nine settings. On M3D/Qwen3-VL-8B, Hulu-Med L1 scored higher while retaining 1.7 times as many visual tokens.

With $\rho_{\max}=0.30$, \method{} retained 22.8\% of visual tokens on average. MedGemma-1.5-4B retained 8.6\%--9.6\% of visual tokens, whereas Hulu-7B and Qwen3-VL-8B operated close to the upper bound. This pattern is consistent with the attention profiles in Fig.~\ref{fig:motivation}. Where realized retention was close to the fixed 10\% setting, the scores were also similar. For MedGemma-1.5-4B, differences from the fixed 10\% results were at most 1.33 across the six reported metrics, all expressed on 0--100 scales.

\begin{table}[!t]
\caption{Component ablations on AMOS-MM at 10\% nominal retention with matched per-study token counts. Full gives the absolute scores of \method{}; the other columns give Full minus each control, which uses a cosine or Euclidean anchor relation, uniform allocation, or no refinement. Positive values favor \method{}.}
\label{tab:components}\centering\tableformat
\begin{tabularx}{\columnwidth}{@{}l*{5}{>{\centering\arraybackslash}X}@{}}
\toprule
& Full & \multicolumn{2}{c}{Anchor relation} & Allocation & Refinement \\
\cmidrule(lr){2-2}\cmidrule(lr){3-4}\cmidrule(lr){5-5}\cmidrule(l){6-6}
Metric & Score & Cosine & Euclidean & Uniform & Removed \\
\midrule
\multicolumn{6}{@{}l}{\textit{Hulu-7B}} \\
ROUGE-L & 37.41 & $+15.42$ & $+17.39$ & $+1.95$ & $+2.21$ \\
METEOR & 28.35 & $+13.02$ & $+13.21$ & $+3.72$ & $+2.15$ \\
\midrule
\multicolumn{6}{@{}l}{\textit{MedGemma-1.5-4B}} \\
ROUGE-L & 17.99 & $-1.67$ & $+1.04$ & $+2.34$ & $+2.15$ \\
METEOR & 10.56 & $-0.70$ & $+2.87$ & $+0.75$ & $+3.76$ \\
\midrule
\multicolumn{6}{@{}l}{\textit{Qwen2.5-VL-7B}} \\
ROUGE-L & 19.64 & $+4.82$ & $+5.45$ & $+3.09$ & $+2.57$ \\
METEOR & 15.46 & $+1.45$ & $+2.56$ & $+2.36$ & $-1.42$ \\
\midrule
\multicolumn{6}{@{}l}{\textit{Qwen3-VL-8B}} \\
ROUGE-L & 19.25 & $+4.07$ & $+5.11$ & $+2.98$ & $+1.82$ \\
METEOR & 17.27 & $+1.49$ & $+1.30$ & $+0.79$ & $-1.37$ \\
\bottomrule
\end{tabularx}
\end{table}

\begin{figure*}[!t]
\captionsetup{hypcap=false}
\begin{minipage}[t]{\columnwidth}
\vspace{0pt}
\centering
\includegraphics[width=\linewidth,height=.926\linewidth,keepaspectratio]{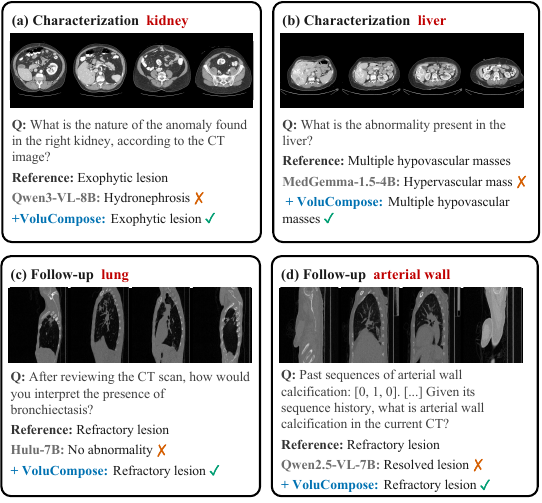}
\caption{Qualitative VQA comparisons on M3D~\cite{m3d2024} and 3D-RAD~\cite{rad2025}. Predictions of the same backbone without compression and with \method{} are compared with reference answers on abnormality morphology, vascular appearance, and current status.}
\label{fig:vqa_cases}
\end{minipage}\hfill
\begin{minipage}[t]{\columnwidth}
\vspace{0pt}
\centering
\includegraphics[width=\linewidth,height=.926\linewidth,keepaspectratio]{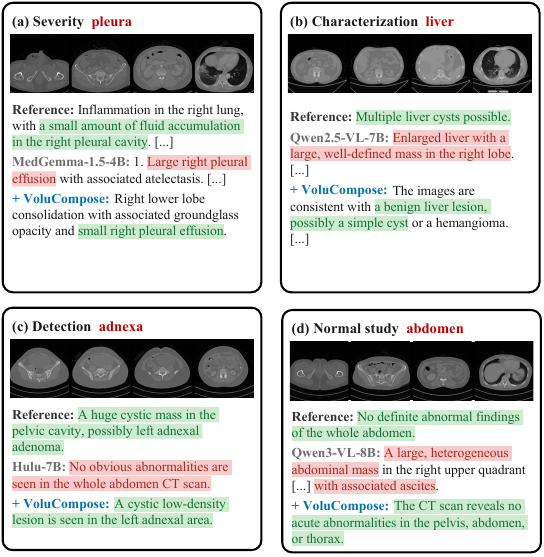}
\caption{Qualitative report-generation comparisons on AMOS-MM~\cite{amos2024}. Reference report excerpts are compared with outputs of the same backbone without compression and with \method{}. Highlighted passages mark differences in localized findings, abnormality attributes, and unsupported descriptions; ellipses indicate omitted slices or text.}
\label{fig:report_cases}
\end{minipage}
\end{figure*}

\subsection{Component Ablations at Matched Token Counts}
\label{sec:components}

Both budget protocols use the same volumetric composition procedure. We examine inter-unit allocation, anchor relations, and local refinement at matched per-study token counts.

On AMOS-MM, four controls are evaluated at a nominal 10\% budget, each matching the number of visual tokens retained by \method{} for every study (Table~\ref{tab:components}). The cosine and Euclidean controls replace the Lorentz distance in anchor ranking and neighborhood construction. The uniform-allocation control distributes the realized token count evenly across the same support units, and the no-refinement control omits the neighborhood update.

Matched-token controls showed complementary, interface-dependent contributions (Table~\ref{tab:components}). Inter-unit allocation improved both metrics over uniform allocation for every backbone. Using the Lorentz distance for anchor ranking and refinement outperformed the Euclidean control for all four backbones and the cosine control for three. Mean ROUGE-L on AMOS-MM was 23.57, 17.91, and 16.32 with the Lorentz distance, cosine similarity, and Euclidean distance, respectively. These comparisons identify the feature relation as a major contributor to report-generation performance. The Euclidean control also retains magnitude information, suggesting a role for the coupling in Eq.~\eqref{eq:coupling}. However, these controls change both anchor ranking and neighborhoods, without isolating exponential gating.

MedGemma-1.5-4B was an exception: the variant with cosine similarity exceeded the full method by 1.67 points in ROUGE-L and 0.70 points in METEOR. Figure~\ref{fig:process_example} illustrates that selected anchors can include background locations. Vision Transformers can assign high feature norms to background tokens~\cite{darcet2024registers}, which may affect magnitude-aware comparisons.

Refinement improved ROUGE-L for all four backbones, with mean gains of 2.19 points in ROUGE-L and 0.78 points in METEOR across backbones. For nonempty neighborhoods, averaging halves each anchor's deviation from its neighborhood mean, balancing anchor fidelity and contextual aggregation (Eq.~\eqref{eq:refinement_objective}). This change in the features supplied to the frozen decoder may affect generated content and wording differently across backbones. Since ROUGE-L and METEOR use different matching and ordering criteria~\cite{rouge2004,meteor2005}, their responses need not be uniform, consistent with the metric differences observed for the Qwen backbones.

\subsection{Qualitative Analysis of Token Composition}
\label{sec:qualitative}
In the Hulu-7B example of Fig.~\ref{fig:process_example}, 24 slices formed 11 ordered support units with nonuniform token allocations. Visually repetitive units received fewer tokens than several subsequent units. Refined anchors incorporated related context from spatially separated patches without adding tokens. Fig.~\ref{fig:semantic_examples} shows that retained locations can coincide with regions of high visual--text correlation despite substantial token reduction. These maps are used for post hoc inspection; neither module uses the question or report text for compression.

Figs.~\ref{fig:vqa_cases} and~\ref{fig:report_cases} present qualitative examples spanning detection, characterization, severity grading, normal-study recognition, and follow-up assessment. In the VQA examples, \method{} distinguished an exophytic renal lesion from hydronephrosis and identified multiple hypovascular hepatic masses, correcting discrepancies in the uncompressed outputs. In the report-generation examples, it recovered an omitted left adnexal cystic lesion, described a right pleural effusion as small rather than large, and avoided unsupported mass and ascites descriptions in a reference-negative case.

These paired cases suggest that consolidating redundant context can improve individual responses. This interpretation is consistent with reported difficulties in using long textual contexts~\cite{liu2024lost}.

\subsection{Analytical Decoder Resource Savings}
\label{sec:cost}
In a separate resource analysis on two VQA datasets and four backbones, the nominal 10\% budget reduced estimated decoder prefill FLOPs by 92.05\% and KV-cache demand by 89.23\%. With a 30\% retention cap, the calibrated protocol yielded reductions of 77.50\% and 69.40\%, respectively. Each setting includes 49 profiles, and each protocol is summarized by the median of its eight setting-level medians. The estimates use total decoder sequence lengths, including visual and nonvisual tokens. Vision encoding and token composition fall outside these decoder estimates.

\section{Conclusion}
\label{sec:conclusion}
We presented \method{}, a training-free, model-agnostic framework that decouples hyperbolic token composition from attention-calibrated capacity control. The framework consolidates volumetric content and supports prescribed or attention-calibrated budgets across heterogeneous visual interfaces. Comparisons under calibrated and prescribed budgets, together with matched-token ablations, supported effective compression across the evaluated backbones. This decoupled design supports efficient full-volume medical image understanding.

\FloatBarrier
\clearpage
\paperreferencebreak{ye2026unveiling}
\bibliographystyle{volucompose}
\bibliography{references}
\end{document}